\documentclass[11pt]{article}

\usepackage[preprint]{acl}
\usepackage{times}
\usepackage{latexsym}

\usepackage[T1]{fontenc}
\usepackage[utf8]{inputenc}
\usepackage{microtype}
\usepackage{inconsolata}
\usepackage{graphicx}
\usepackage{amssymb}
\usepackage{pifont}
\usepackage{threeparttable}

\usepackage{booktabs,multicol, multirow}
\usepackage{microtype}
\usepackage{hyperref}
\usepackage{url}
\usepackage{booktabs}
\usepackage{caption}
\usepackage{quoting}
\usepackage{inconsolata}
\usepackage{url}
\usepackage{graphicx}
\usepackage{xspace}
\usepackage[subtle]{savetrees}
\usepackage[most]{tcolorbox}
\usepackage{colortbl}
\usepackage{microtype}      
\usepackage{xcolor}
\usepackage{multirow}
\usepackage{multicol}
\usepackage{longtable}
\usepackage{listings}
\usepackage{enumitem}
\usepackage{psfrag}
\usepackage{verbatim}
\usepackage{pifont}
\usepackage{cleveref}
\usepackage{adjustbox}

\newcommand{\OURS}{\texttt{DeepBind}\xspace}
\newcommand{\BASEQA}{\texttt{QA}\xspace}
\newcommand{\BASEBIO}{\texttt{Bio}\xspace}
\newcommand{\BASEPORTRAY}{\texttt{Portray}\xspace}

\newcommand{\DATE}{\textit{Temporal Grounding}\xspace}
\newcommand{\CONSISTENCY}{\textit{Consistency Filtering}\xspace}

\newcommand{\sig}[1]{%
  \ifnum#1=1 $^{*}$%
  \else\ifnum#1=2 $^{**}$%
  \else\ifnum#1=3 $^{***}$%
  \else%
  \fi\fi\fi%
}

\definecolor{darkblue}{rgb}{0, 0, 0.5}
\hypersetup{colorlinks=true, citecolor=darkblue, linkcolor=darkblue, urlcolor=darkblue}

\definecolor{berkeleyblue}{HTML}{3B7EA1}
\definecolor{berkeleygold}{HTML}{FDB515}
\definecolor{customgray}{HTML}{888888}
\definecolor{darkgray}{HTML}{222222}
\definecolor{main}{HTML}{4472C4}   
\definecolor{sub}{HTML}{EBF4FF} 
\definecolor{github}{HTML}{181717}
\definecolor{hf}{HTML}{FC9313}

\definecolor{berkeleyblue}{HTML}{3B7EA1}
\definecolor{berkeleygold}{HTML}{FDB515}
\definecolor{customgray}{HTML}{888888}
\definecolor{darkgray}{HTML}{222222}
\definecolor{main}{HTML}{4472C4}   
\definecolor{sub}{HTML}{EBF4FF}

\newcommand\hy{\rowcolor{berkeleygold!20}}

\newcommand\hg{\rowcolor{customgray!20}}

\title{Identity, Cooperation and Framing Effects 

within Groups of Real and Simulated Humans}

\author{Suhong Moon\textsuperscript{1}\thanks{Equal contribution}, Minwoo Kang\textsuperscript{1$*$}, 
Joseph Suh\textsuperscript{1}, 
Mustafa Safdari\textsuperscript{2}\thanks{No Google-affiliated authors used or analyzed any large language models, including LLaMA in this work.}, 
John Canny\textsuperscript{1,2}\footnotemark[2] \\
  \textsuperscript{1}University of California, Berkeley,
    \textsuperscript{2} Google DeepMind \\
  \texttt{suhong.moon@berkeley.edu}}

\begin{document}
\maketitle

\begin{abstract}
Humans act via a nuanced process that depends both on rational deliberation and also identity and contextual factors.
In this work, we study how large language models (LLMs) can simulate  human action in the context of social dilemma games. 
While prior work has focused on ``steering'' (weak binding) of chat models to simulate personas, we analyze here how deep binding of base models with extended backstories
leads to more faithful replication of
identity-based behaviors. 
Our study has these findings: simulation fidelity vs human studies is improved by conditioning base LMs with rich context of narrative identities and checking consistency using instruction-tuned models. 
We show that LLMs can also model contextual factors such as time (year that a study was performed), question framing, and participant pool effects. LLMs therefore allow us to explore the details that affect human
studies but which are often omitted from 
experiment descriptions, and which hamper accurate replication. 
\end{abstract}
\section{Introduction}
\begin{figure}[t]
    \centering
    \includegraphics[width=\linewidth]{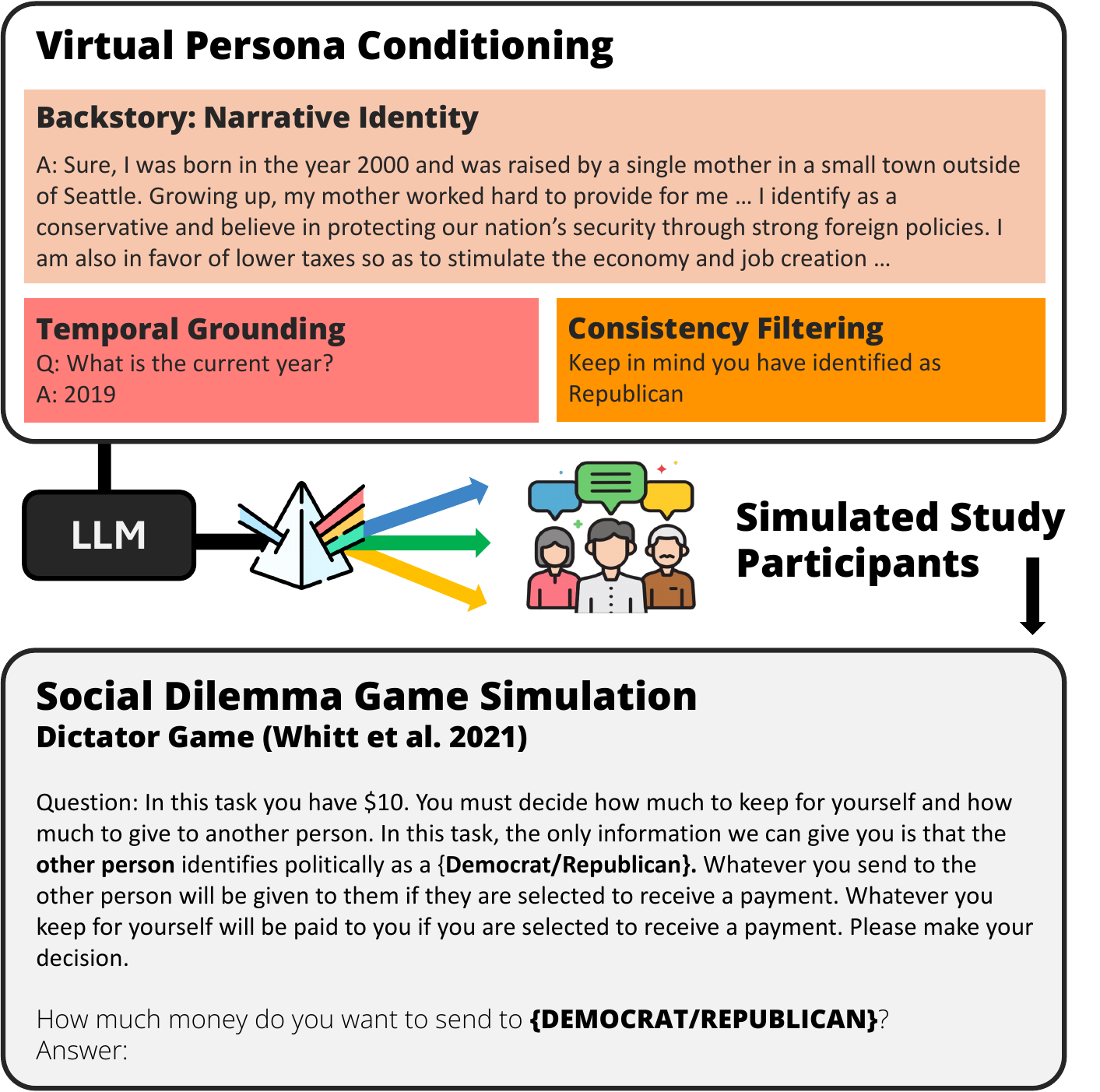}
    \caption{
        \small{
            \textbf{Deep Persona Binding and Consistency Filtering.}
In this work, pretrained language models are deeply bound with extended backstories, then \DATE is applied and \CONSISTENCY is used to remove implausible or inappropriate responses. The resulting human model more accurately predicts identity driven behaviors and other contextual effects.
        }
    }
    \label{fig:intro_questions}
    \vspace{-10pt}
\end{figure}
Human decision making is a complex process shaped not only by rational deliberation but also by social identity and other contextual factors.
Decades of research in social psychology and political science have shown that individuals cooperate better with ingroup peers than with outgroups~\cite{tajfel1986social,tajfel1979integrative,iyengar2012affect,iyengar2015fear,whitt2021tribalism,fowler2007beyond}.
These dynamics become especially salient in social dilemma games such as the Dictator and Trust Games, where choices directly reveal altruism, trust, and perceptions of reciprocity. 
Recent studies demonstrate that partisan identity can exert stronger effects than race on patterns of generosity and cooperation~\cite{iyengar2012affect}.

While recent large language models (LLMs) have emerged as promising tools for simulating human behavior~\cite{park2022social,kang2025higher,abdulhai2023moral,park2023generative,moon2024virtual,suh2025language}, prior work has been limited in critical ways. Most existing research has focused on opinion/attitude surveys rather than behavior~\cite{park2024generative, argyle2023out, santurkar2023whose,moon2024virtual}. Work which has considered behavior has focused on rational decision making without considering identity effects~\cite{xie2024trust,aher2023using,fan2024can,light2023avalonbench}. This leaves a significant gap: can LLMs simulate not just what people say, but what they actually do—particularly when those actions are influenced by social identity? 

Predicting behavior in realistic social contexts presents a fundamentally different challenge than replicating survey responses. Behavioral studies require models to generate decisions that simultaneously reflect (1) individual identity characteristics, (2) perceptions of group membership (both their own and others'), and (3) context-dependent strategic reasoning about cooperation and reciprocity. 
In particular, partisan animosity has grown steadily from the mid-2000s to the mid-2020s, and various studies have shown stronger effect sizes for
partisan disparities in recent years~\citep{whitt2021tribalism,iyengar2012affect,iyengar2015fear,carlin2018political}. So we explore the date of the study as an additional contextual factor in this work.

In this paper, we explore how well LLMs can model human behavior including identity and temporal context. Specifically, we explore how well LLM virtual personas can reproduce co-partisan favoritism in resource allocation and trust decisions. 
We introduce two prompting strategies: \CONSISTENCY, which reiterates persona identity and group membership throughout long simulations, and \DATE, which situates simulations in the year corresponding to the original human study. 
We evaluate these methods using backstory-conditioned pretrained models, which remains the most accurate zero-shot simulation method~\cite{moon2024virtual,kang2025higher,vezhnevets2023generative}.

Our results show that deep backstory binding, temporal grounding and consistency checking improve alignment between LLM-simulated actions and human empirical data. 
In the present case, it allows us to \emph{explore the disparities in results} between human replication studies of the same phenomenon but with subtly different experiment design. We can compare effects of obvious differences like the year of study, and more subtle differences like the wording of experiment instructions, and participant pool effects. 
\section{Related Work}
\subsection{LLMs for Simulating Human Behavior}
Recent research has highlighted the potential of LLMs for \emph{simulating} humans over a range of topics, from political science~\citep{jiang2022communitylm,simmons2022moral,hartmann2023political,wu2023large,kim2023ai,bail2023we,bail2024can,chu2023language}, economics~\citep{Fatouros2024CanLL,phelps2023models,horton2023large}, to psychology~\citep{karra2022estimating,perez2023discovering,binz2023using,jiang2023personallm,serapio2023personality,hilliard2024eliciting,suh2024rediscovering,qian2025mask}.  
By conditioning to virtual personas, these models can mimic patterns of human reasoning and social interaction across diverse contexts~\cite{park2023generative, park2024generative}.
Most studies have focused on surveys, notably cross-sectional opinion polls~\cite{santurkar2023whose, argyle2023out, park2024generative, moon2024virtual, suh2025language} or have
neglected identity effects. 
Instead, our present work explores LLM-based simulation to \emph{action prediction with contextual factors} like identity, framing and study date.

\subsection{Behavioral Games with LLMs}

LLM-based simulated humans have been tested in classic game-theoretic scenarios.
For example, \citet{lan2023llm,wang2023avalon,light2023avalonbench,shi2023cooperation} investigated LLM cooperation in games and showed that models can qualitatively reproduce human-like decision traits  such as reciprocity and risk-aversion consistent with status quo bias ~\cite{horton2023large, xie2024trust, Murashige2025Ultimatum}.
However prior work has focused on rational behavior or behavioral biases, and not only on group identity or other social factors.
\section{Why Pretrained LLMs ?}
\label{sec:axelrod_why_pretrained_llm}
Here we discuss our rationale for simulating users with pretrained language models (aka \emph{base} models) rather than fine-tuned chat models. To be precise, a pretrained model is the first stage of modern LLM training, where the model is trained with next-token or masked token loss on an enormous corpus of training data, typically terabytes, including a significant amount of human dialog data. It is the most expensive step in training current models. 
Prior studies~\cite{kang2025higher,moon2024virtual,suh2025language,santurkar2023whose} have shown that carefully-prompted pretrained models outperform instruction-tuned variants in human simulation tasks.
But it is not just a question of accuracy but of \emph{fair representation}. Instruction-tuning \emph{biases models away from fair representation} of people, as we shall see in a moment. 

\begin{figure}[t]
    \centering
    \includegraphics[width=\linewidth]{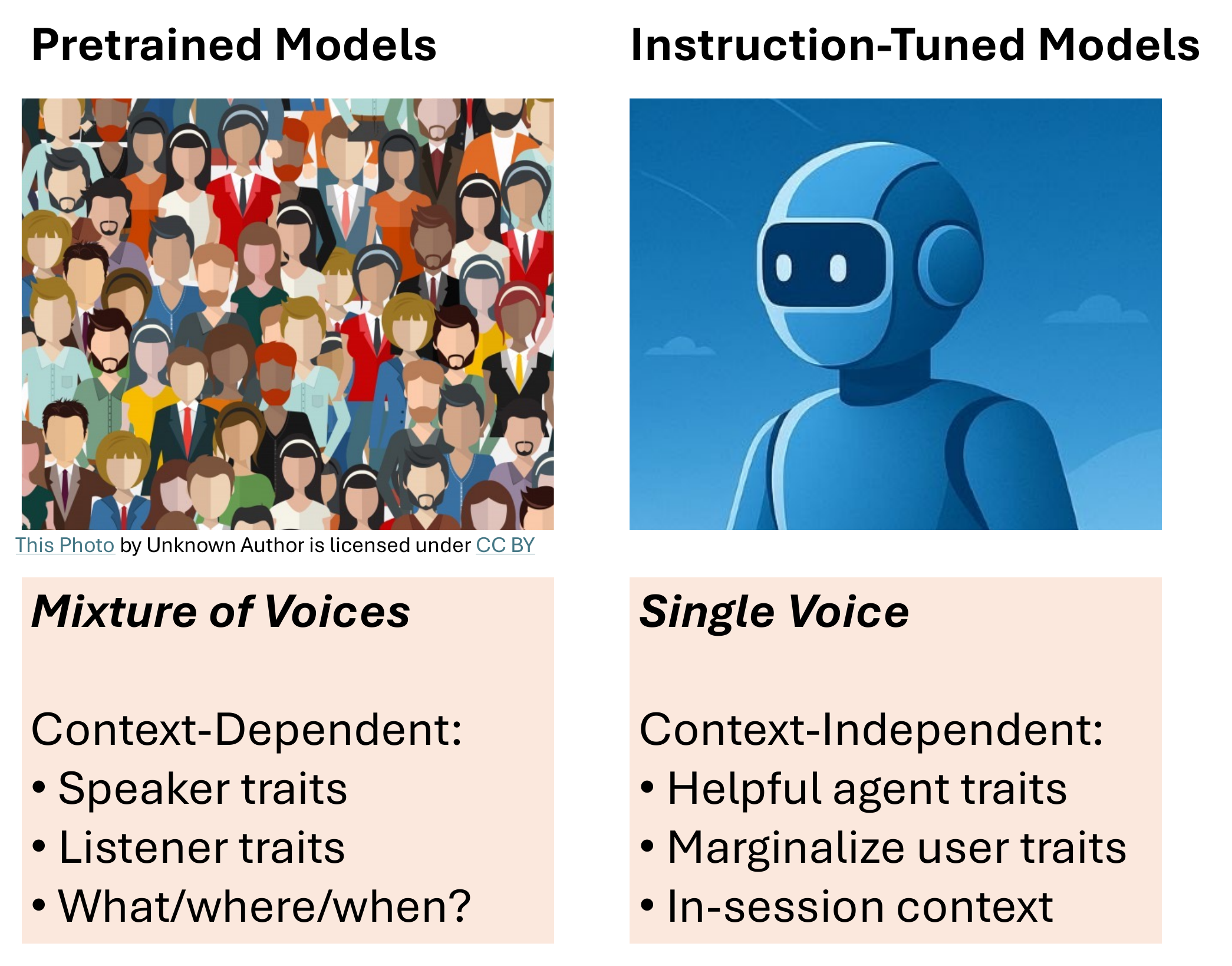}
    \caption{
        \small{
            \textbf{Conceptual illustration of the differences between pretrained and fine-tuned language models.}
        }
    }
    \label{fig:axelrod_pt_it_models}
    
\end{figure}

First of all, it's important to understand that pretrained models are not ``agent models'' in the same way that instruction-tuned models are, see \Cref{fig:axelrod_pt_it_models}. Pretrained model data consist of billions of text snippets with latent (hidden) context information such as the speaker, listener and other contextual factors. They therefore exactly model the ``voices'' of billions of users in billions of contexts. 

To ``bind'' or ``steer'' a pretrained model to a particular speaker and context, it suffices to provide sufficient context in a text prompt, and the model will faithfully extend the text from that speaker/context. There are some challenges in doing this reliably, but we address those in a moment. But its important at this stage to reflect on how well pretrained models fit to the task of simulating an enormously diverse population of people. If one were to design a human simulation model from scratch it would arguably be a pretrained model. 

But the natural context-dependence of pretrained models, an asset for human simulation, is a liability in agent applications. Unless the speaker, listener and context are fully-constrained, the LLM will sample from a posterior of contexts that fit the text prompt so far, and give different responses based on that sampling. After pretraining, LLM training comprises several QA-prompt optimization stages that remove the pretrained model's context dependence, freezing the model's identity to a ``helpful agent'' persona, the user's identity to ``general user'', and the context to whatever can be inferred
within the current chat session. The model is intentionally steered away from diversity in agent personality, emotion, complex social goals etc, so as to maximize the model's repeatability on a given prompt.

\begin{figure}[t]
    \centering
    \includegraphics[width=\linewidth]{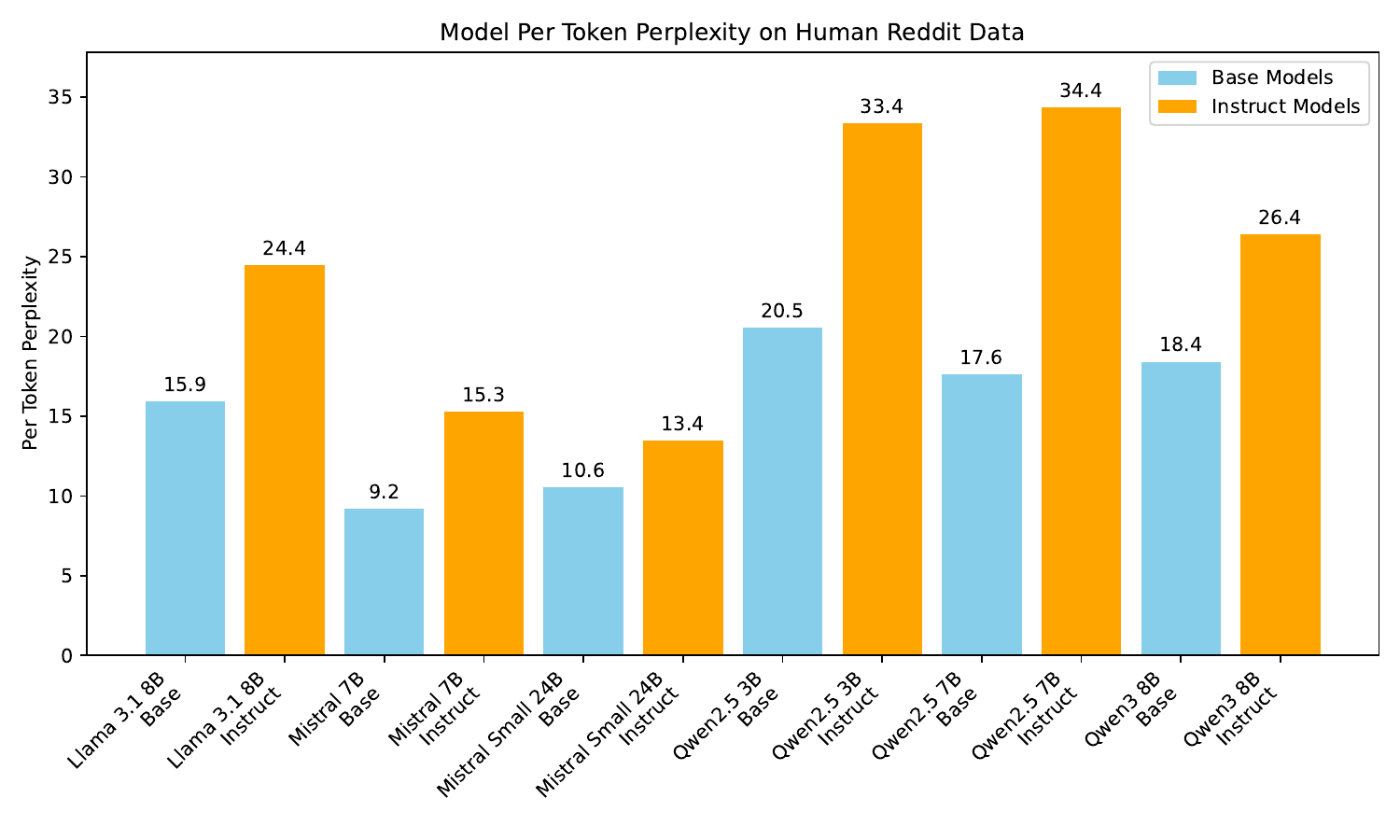}
    \caption{
        \small{
            \textbf{Per-token perplexity of base and instruction-tuned models on human Reddit data.} 
            Instruction-tuned models (orange on right) exhibit substantially higher perplexity across all model families, compared to same-sized pretrained variants (blue on left).
        }
    }
    \label{fig:axelrod_model_perplexity}
    \vspace{-10pt}
\end{figure}

The effects of this later IT model training can be seen in \Cref{fig:axelrod_model_perplexity}. By their nature, post-training datasets are much smaller than pretraining data. It is not possible to preserve the high accuracy of PT models on human dialog during later training, and this can be seen in the difference in perplexity between various base models and their IT variants on human dialog data. Typical increases are 50\% to 100\% for the IT models. Note also that the best pretrained models have single-digit perplexities, which is extraordinarily low for English text. 

Finally the effects of IT model tuning are not uniformly distributed across personal traits, but rather weaken or eliminate negative human traits, and thereby under-represent individuals that have them. In fact, the effect is so strong that it is likely that \emph{most users} are not represented. We call this the {\bf Lake Wobegon Effect} (everyone is above human average in the resulting IT-model cohort). 

For example, in ~\cite{li2025llmgeneratedpersonapromise} the authors found that for IT-model generated virtual personas ``Notably
absent are terms reflecting life challenges, social difficulties, or negative experiences, suggesting
LLMs may be systematically avoiding less favorable characterizations.'' The authors then go on to describe a variety of biases in results generated by IT models on these personas, concluding that such simulations are almost never in agreement with human studies. Like many recent works, the authors attribute these biases to LLMs in general, rather than the (expected) biases from instruction-tuning. The authors rightfully point out the dangers of using the outcomes from such simulations for decision making, but over-attribute those dangers to all LLMs. Rather, their results are an argument against simulation with IT models. Such biases are not shared by PT models, which instead capture an extraordinarily diverse collection of human traits. 

\section{Identity Binding in LLM Personas}
\label{sec:method}
We have seen that PT models are naturally contextualizable by prompting, but such prompts must be sufficiently detailed to 
capture an individual human ``voice''. 
We use the ``anthology'' approach described in ~\citet{moon2024virtual} and evolved in \citet{kang2025higher}. In particular \citet{kang2025higher} showed that longer backstories create deeper persona binding and improved accuracy in a social attitude test. 

\subsection{LLM Personas via Narrative Backstories}

In this work, each persona backstory is generated by interviewing a PT model. 
Samples of the backstories used in this experiment are provided in \Cref{asec:backstories}.
The interview questions are drawn from the American Voices Project~\citep{stanford2021american}. 
For accurate text generation relative to PT model training data, the models are run at the temperature of $1.0$. 
However, not \emph{all} PT model training data is human dialog, and neither is PT-model generated output. So we rejection-sample non-dialog results using an instruction-tuned critic.
These include code blocks, markup, factual inconsistencies, and non-dialog text structure, similar to \citet{kang2025higher}.
Demographic traits of each virtual persona are then
extracted from their profile using multi-choice questions. 

After creating the backstories and demographics, individual human participant demographics (which were kindly provided by the authors of the human studies below), are matched against all virtual participants to find the demographically closest virtual participant. 
This pool of matching virtual participants are given the same text prompts as were the human participants. 
For more detail of matching process, please refer to \Cref{asec:demographic_matching}.

\subsection{Temporal Grounding and Consistency Filtering}

We add two prompting methods—\textbf{Temporal Grounding} and \textbf{Consistency Filtering}—to improve the fidelity and stability of simulations.

\DATE situates each simulation in the year the corresponding human study was conducted. 
The date is bound to the persona with 
a forced QA pair ``Interviewer: What Year is it?'', ``Model: 2007''.
We were motivated to study temporal effects by~\citet{whitt2021tribalism}, which obtained larger partisan effects than ~\citet{iyengar2015fear}, and conjectured that they were due to a increase in partisanship in the US over that time. 

\CONSISTENCY is used to mitigate a variety of flaws in pretrained LLM output. This includes semantic drift because of short training segment lengths, but also the out-of-dialog genre cases mentioned earlier. However \CONSISTENCY prompts must be designed and used with care. Real humans are often inconsistent, e.g. because of mental health issues, dementia, deception etc.

\section{Can LLMs Simulate In-/Out-group Biases in Decision Making?}
\label{sec:axelrod_experiment}
We explored two widely used behavioral economics games: the Dictator Game and the Trust Game. 
These games have been used in prior political science research to uncover affective polarization, or the tendency of partisans to exhibit more favorable behavior toward in-group (co-partisan) members and less favorable behavior toward out-group (rival partisan) members.

\subsection{Decision Asymmetry in Dictator Games}
\paragraph{Dictator Game.} In this game, one participant (the “dictator”) decides how to divide a fixed amount of money between themself and another participant. 
Prior studies~\cite{fowler2007beyond,iyengar2015fear,whitt2021tribalism} show that partisans are significantly more generous toward co-partisans than rival partisans. 
For example, \citet{fowler2007beyond} find that participants allocate more to co-partisans than to opposing partisans in dictator games, even at personal cost.
Both altruism and identity-based giving are linked to higher political participation.

\paragraph{Experimental Setup and Metrics.}
Our metrics are based on those in~\citet{iyengar2015fear,whitt2021tribalism}. 
We define the in-/out-group trust asymmetry in dictator games as the difference in mean transfer amounts allocated to co-partisans versus rival partisans.

Details of the prompts for \citet{iyengar2015fear} and \citet{whitt2021tribalism} are included in \Cref{asec:study_instructions}. We also prompted for the year as outlined above.

\paragraph{Baselines.}
We adopt the \BASEQA, \BASEBIO, and \BASEPORTRAY prompting strategies from~\citet{santurkar2023whose} as baselines. Each method conditions the model using demographic attributes such as age, gender, race, education, income, political affiliation, and other relevant factors.

\begin{itemize}
\vspace{-0.5em}
\item \BASEQA presents a series of question–answer pairs for each demographic variable (e.g., \textit{Q: What is your political affiliation? A: Republican}). Both Q and A texts are forced and presented to the model with the answer in the model's first-person name. 
\vspace{-0.5em}
\item \BASEBIO generates rule-based free-text biographies written in the first person (e.g., \textit{I am a Republican}).
\vspace{-0.5em}
\item \BASEPORTRAY produces similar biographies but written in the second person (e.g., \textit{You are a Republican}).
\end{itemize}

\subsection{Decision Asymmetry in Trust Games}
\paragraph{Trust Game.} This game involves two players: a sender and a receiver. The sender decides how much money to send to the receiver, with the amount tripled in transit; the receiver then decides how much to return. This game measures both trust (by the sender) and trustworthiness (by the receiver). Studies show that senders are more likely to trust co-partisans, and expect more generous returns from them~\cite{whitt2021tribalism}.

\paragraph{Experimental Setup and Metrics.}
We use the measures from prior human studies~\cite{carlin2018political, whitt2021tribalism}.
The main difference from the dictator game is that the Trust Game involves both the sender's disposition toward the receiver, and also their belief about the receiver reciprocating. 
This makes the task more complex: models must capture both ingroup trust (greater willingness to send money to co-partisans) and ingroup expectations (belief that co-partisans will return more).

\begin{table*}[h]
    \centering
    \caption{
        \textbf{Simulating Partisan Bias in Dictator and Trust Games~\citep{iyengar2015fear,whitt2021tribalism,carlin2018political}.}
        Dem $\Delta$ measures the difference in the amount of money Democratic participants allocate to co-partisans versus opposing partisans.  
        Similarly, Rep $\Delta$ captures the same difference for Republican participants.  
        The best-performing method for each model is shown in \textbf{bold}, and the overall best for each column is highlighted with \underline{underline}.
    }
    \label{tab:main_results}
    \resizebox{0.9\textwidth}{!}{
        \begin{tabular}{cc|cc|cc|cc|cc}
            \toprule
            \multirow{3}{*}{Model} & \multirow{3}{*}{Method} 
            & \multicolumn{4}{c|}{Dictator Game} 
            & \multicolumn{4}{c}{Trust Game} \\
            & & \multicolumn{2}{c|}{\small \citet{iyengar2015fear}} & \multicolumn{2}{c|}{\small \citet{whitt2021tribalism}} 
              & \multicolumn{2}{c|}{\small \citet{carlin2018political}} & \multicolumn{2}{c}{\small \citet{whitt2021tribalism}} \\
            & & Dem $\Delta$ & Rep $\Delta$ & Dem $\Delta$ & Rep $\Delta$
              & Dem $\Delta$ & Rep $\Delta$ & Dem $\Delta$ & Rep $\Delta$ \\
            \midrule
            \multirow{4}{*}{Mistral-Small (24B)} 
                & \BASEQA                   & 0.79 & 0.15 & 2.21 & 1.71 & 0.91  & \textbf{1.17} & 1.19 & 1.06 \\
                & \BASEBIO                  & 1.05 & 0.11 & 2.00 & 1.46 & 1.07  & 0.43 & 2.20 & 1.28 \\
                & \BASEPORTRAY              & 0.50 & 0.30 & 2.14 & 1.52 & 0.23 & \underline{1.14} & 1.86 & 0.88 \\
            \hy \cellcolor{white} & \OURS   & 0.86 & 0.58 & \underline{2.60} & \underline{2.16} & 0.83  & 1.37 & 1.49 & \textbf{2.00} \\
            \midrule
            \multirow{4}{*}{Mixtral 8x22B} 
                & \BASEQA                   & 0.26 & 0.22 & 2.15 & 1.87 & 1.07 & 0.99 & 1.92 & 0.54 \\
                & \BASEBIO                  & 1.04 & 0.18 & 1.83 & 1.65 & 0.02 & 1.41 & 2.21 & 0.53 \\
                & \BASEPORTRAY              & 0.28 & 0.24 & 1.83 & 1.70 & 0.56 & 0.86 & 1.41 & 0.95 \\
            \hy \cellcolor{white} & \OURS   & \textbf{0.70} & \textbf{0.60} & 2.62 & \textbf{2.37} & \textbf{0.65} & 1.04 & \underline{2.15} & 1.31 \\
            \midrule
            \multirow{4}{*}{Qwen-2.5 72B} 
                & \BASEQA                   & 1.66 & 0.26 & 2.14 & 1.60 & 0.96 & 0.57 & 1.51 & 0.65 \\
                & \BASEBIO                  & 0.89 & 0.29 & 1.83 & 1.23 & 0.47 & 0.43 & 1.60 & 1.13 \\
                & \BASEPORTRAY              & 1.07 & 0.12 & 1.77 & 1.42 & 0.40 & 0.72 & 1.79 & 1.08 \\
            \hy \cellcolor{white} & \OURS   & \underline{0.72} & \underline{0.73} & \textbf{2.58} & 1.87 & \underline{0.60} & 0.97 & \textbf{2.14} & \underline{1.38} \\
            \midrule
            \hg \multicolumn{2}{c|}{Human} 
                & 0.66 & 0.65 & 2.46 & 2.33 & 0.65 & 1.22 & 2.11 & 1.76 \\
            \bottomrule
        \end{tabular}
    }
    \vspace{0.5em}
\end{table*}

We use the same persona conditioning baselines as in the dictator game—\BASEQA, \BASEBIO, \BASEPORTRAY—as well as our proposed \OURS method.
For each method, we compute $\Delta$ metrics separately for Democratic and Republican senders, measuring the difference in the amount sent to co-partisans versus opposing partisans. 
Higher $\Delta$ values indicate stronger partisan bias in trust behavior. Our evaluation focuses on how well the models replicate these empirically observed asymmetries in trust across party lines.
\section{Experimental Results}
\subsection{Comparison with Baselines}
\Cref{tab:main_results} reports the simulated partisan bias observed in Dictator and Trust Games across three large language models—Mistral-Small (24B), Mixtral 8x22B, and Qwen-2.5 72B ~\citep{mistral2025small24b,mixtral-8x22b,qwen2.5}—under various persona conditioning methods. 
We use both \CONSISTENCY and \DATE for all cases.
We compare four prompting strategies: \BASEQA, \BASEBIO, \BASEPORTRAY, and our method, labeled \OURS.

\begin{table*}[!h]
    \centering
    \caption{
        \textbf{Ablation: Effects of Socio-Temporal Grounding in Simulating Dictator Games~\citep{iyengar2015fear, whitt2021tribalism}}  
        Results from replicating human actions in dictator games, where human participants exhibit favoritism towards co-partisan recipients.
    }
    \label{tab:main_results_dictator_game}
    \resizebox{\textwidth}{!}{
    \begin{tabular}{cc|ccc|ccc|c}
    \toprule
    Model & Method & Dem$\rightarrow$Dem & Dem$\rightarrow$Rep & Dem $\Delta$ & Rep$\rightarrow$Rep & Rep$\rightarrow$Dem & Rep $\Delta$ & Mean Diff.\ $\Delta$ \\
    \midrule
    \multirow{4}{*}{Mistral-Small 24B} & No date, No consistency & 4.62 & 3.96 & 0.66 & 4.01 & 4.01 & 0.00 & 0.33 \\
                                   & Consistency              & 4.55 & 3.97 & 0.58 & 4.13 & 3.90 & 0.23 & 0.41 \\
                                   & Date 2014                & 4.35 & 4.17 & 0.18 & 4.28 & 4.19 & 0.09 & 0.14 \\
       \hy\cellcolor{white}        & Date 2014 + Consistency  & 4.56 & 3.70 & 0.86 & 3.95 & 3.38 & 0.58 & 0.72 \\
    \midrule
    \hg \multicolumn{2}{c|}{Human 2014 \cite{iyengar2015fear}}  & 3.82 & 3.14 & 0.68 & 3.21 & 2.56 & 0.65 & 0.67 \\
    \bottomrule
\end{tabular}
    }
    \vspace{0.5em} 
    
    \resizebox{\textwidth}{!}{
    \begin{tabular}{cc|ccc|ccc|c}
    \toprule
    Model & Method & Dem$\rightarrow$Dem & Dem$\rightarrow$Rep & Dem $\Delta$ & Rep$\rightarrow$Rep & Rep$\rightarrow$Dem & Rep $\Delta$ & Mean Diff.\ $\Delta$ \\
    \midrule
    \multirow{4}{*}{Mistral-Small 24B} & No date, No consistency & 4.13 & 2.84 & 1.29 & 3.70 & 2.70 & 0.99 & 1.14 \\
                                   & Consistency              & 4.20 & 2.36 & 1.84 & 4.14 & 2.74 & 1.40 & 1.62 \\
                                   & Date 2019                & 4.06 & 2.56 & 1.50 & 3.59 & 2.59 & 1.00 & 1.25 \\
       \hy\cellcolor{white}        & Date 2019 + Consistency  & 4.79 & 2.63 & 2.16 & 4.83 & 2.23 & 2.60 & 2.38\\
    \midrule
    \hg \multicolumn{2}{c|}{Human 2019 \cite{whitt2021tribalism}} & 5.37 & 2.91 & 2.46 & 4.68 & 2.35 & 2.33 & 2.40 \\
    \bottomrule
\end{tabular}

    }
\end{table*}

Across all models and games, the \OURS method consistently produces simulated behavior that more closely replicates human partisan gaps ($\Delta$ values). 
In the Dictator Game, \OURS achieves the highest alignment with human behavior for both Democrats and Republicans in most cases. 
For example, in the 2019 study by~\citet{whitt2021tribalism}, 
\OURS yields the highest Republican $\Delta$ for all models and the highest Democratic $\Delta$ for Mixtral and Qwen. 
In the 2014 study by~\citet{iyengar2015fear}, Qwen with \OURS nearly matches the human-level Democratic $\Delta$ (0.72 vs. 0.66), while Mixtral with \OURS performs best among its peers (0.70).

In the Trust Game,  Mistral with \OURS achieves the highest Republican $\Delta$ (2.00), closely matching the human value (1.76). On Mixtral, \OURS produces the highest Democratic $\Delta$ (0.65), which is identical to the human benchmark. Qwen with \OURS similarly produces strong $\Delta$ values, including the highest Republican $\Delta$ (1.38) across all models for the 2019 Trust Game.

Overall, \OURS improves simulation accuracy of partisan bias across both ideological groups and experimental settings. 
While other methods may occasionally outperform \OURS on isolated metrics, \OURS achieves the most consistent alignment with human partisan behavior across all four studies and both games.

\begin{table*}[!t]
    \centering
    \caption{
        \textbf{Ablation: Effects of Socio-Temporal Grounding in Simulating Trust Games~\citep{carlin2018political, whitt2021tribalism}}  
            Results from replicating human actions in trust games, where human participants exhibit favoritism towards co-partisan recipients. 
    }
    \label{tab:main_results_trust_game}
    \resizebox{\textwidth}{!}{
    \begin{tabular}{cc|ccc|ccc|c}
    \toprule
    Model & Method & Dem$\rightarrow$Dem & Dem$\rightarrow$Rep & Dem $\Delta$ & Rep$\rightarrow$Rep & Rep$\rightarrow$Dem & Rep $\Delta$ & Avg.\ $\Delta$ \\
    \midrule
    \multirow{4}{*}{Mistral-Small} & No date, No consistency & 4.01 & 3.98 & 0.04 & 3.63 & 3.58 & 0.05 & 0.04 \\
                                   & Consistency              & 3.83 & 3.77 & 0.06 & 4.27 & 3.66 & 0.61 & 0.33 \\
                                   & Date 2015                & 4.16 & 3.32 & 0.83 & 3.61 & 3.76 & -0.15 & 0.34 \\
       \hy\cellcolor{white}        & Date 2015 + Consistency  & 4.20 & 3.70 & 0.50 & 4.49 & 3.13 & 1.37 & 0.94 \\
    \midrule
    \hg \multicolumn{2}{c|}{Human 2015 \cite{carlin2018political}} & 4.29 & 3.64 & 0.65 & 4.40 & 3.18 & 1.22 & 0.93 \\
    \bottomrule
\end{tabular}

    }
    \vspace{0.5em} 
    
    \resizebox{\textwidth}{!}{
    \begin{tabular}{cc|ccc|ccc|c}
    \toprule
    Model & Method & Dem$\rightarrow$Dem & Dem$\rightarrow$Rep & Dem $\Delta$ & Rep$\rightarrow$Rep & Rep$\rightarrow$Dem & Rep $\Delta$ & Avg.\ $\Delta$ \\
    \midrule
    \multirow{4}{*}{Mistral-Small} & No date, No consistency & 5.28 & 3.60 & 1.68 & 4.79 & 4.23 & 0.56 & 1.12 \\
                                   & Consistency              & 5.12 & 3.59 & 1.53 & 5.08 & 3.83 & 1.25 & 1.39 \\
                                   & Date 2019                & 4.97 & 3.87 & 1.10 & 5.07 & 4.00 & 1.06 & 1.08 \\
       \hy\cellcolor{white}        & Date 2019 + Consistency  & 5.02 & 3.53 & 1.49 & 4.97 & 2.97 & 2.00 & 1.75 \\
    \midrule
    \hg \multicolumn{2}{c|}{Human 2019 \cite{whitt2021tribalism}} & 4.99 & 2.88 & 2.11 & 4.65 & 2.89 & 1.76 & 1.94 \\
    \bottomrule
\end{tabular}
    }
\end{table*}

\begin{table*}[h!]
\centering
\scriptsize
    \caption{
    \textbf{Counterfactual Simulations: Alternative Combinations of Date, Framing, and Participant Pool.}
    Each panel reports all combinations of: study year, framing text, and participant pool. Column labels indicate the source studies:
\textbf{ID} = Iyengar \& Westwood (Dictator Game),  
\textbf{WD} = Whitt et al.\ (Dictator Game),  
\textbf{CT} = Carlin \& Love (Trust Game),  
\textbf{WT} = Whitt et al.\ (Trust Game).   The gray rows  show the original
human experimental results. The orange highlighted
rows reproduce the
design of the adjacent human study. Other rows are counterfactual.
Counter-factual study results are based on the Mistral-Small (24B) pretrained, base model.
    }
    \vspace{-5pt}
    \label{table:axelrod_framing_effects}
    \resizebox{0.75\textwidth}{!}{
        \begin{tabular}{c|c|c|c|c|c|ccc}
        \toprule
        \multicolumn{6}{c|}{\textbf{Counterfactuals}} & \multicolumn{3}{c}{\textbf{Partisan Bias}} \\
        \midrule
        \multicolumn{2}{c|}{Participant Pool} & \multicolumn{2}{c|}{Framing}
        & \multicolumn{2}{c|}{Year}
        & \multirow{2}{*}{Dem $\Delta$}
        & \multirow{2}{*}{Rep $\Delta$}
        & \multirow{2}{*}{Avg $\Delta$} \\
        \cmidrule{1-6}
        ID & WD & ID & WD & ID (2014) & WD (2019) & & & \\
        \midrule
        \hg\multicolumn{6}{c|}{\textbf{Human Study ID \cite{iyengar2015fear}}} & \textbf{0.68} & \textbf{0.65} & \textbf{0.67} \\
        \midrule 
        \hy \checkmark &  & \checkmark &  & \checkmark &  & 0.86 & 0.58 & 0.72 \\
        \checkmark &  & \checkmark &  &  & \checkmark & 1.09 & 1.15 & 1.12 \\
        \checkmark &  &  & \checkmark & \checkmark &  & 1.58 & 1.96 & 1.77 \\
        \checkmark &  &  & \checkmark &  & \checkmark & 1.84 & 2.75 & 2.29 \\
         & \checkmark & \checkmark &  & \checkmark &  & 1.05 & 0.52 & 0.79 \\
         & \checkmark & \checkmark &  &  & \checkmark & 1.15 & 0.59 & 0.87 \\
         & \checkmark &  & \checkmark & \checkmark &  & 1.67 & 1.88 & 1.78 \\
        \hy & \checkmark &  & \checkmark &  & \checkmark & 2.16 & 2.60 & 2.38 \\
        \midrule
        \hg\multicolumn{6}{c|}{\textbf{Human Study WD \cite{whitt2021tribalism}}} & \textbf{2.46} & \textbf{2.33} & \textbf{2.40} \\
        \bottomrule
        \end{tabular}
    }

    \vspace{0.5em}
    
    \resizebox{0.75\textwidth}{!}{
        \begin{tabular}{c|c|c|c|c|c|ccc}
        \toprule
        \multicolumn{6}{c|}{\textbf{Counterfactuals}} & \multicolumn{3}{c}{\textbf{Partisan Bias}} \\
        \midrule
        \multicolumn{2}{c|}{Participant Pool} & \multicolumn{2}{c|}{Framing}
        & \multicolumn{2}{c|}{Year}
        & \multirow{2}{*}{Dem $\Delta$}
        & \multirow{2}{*}{Rep $\Delta$}
        & \multirow{2}{*}{Avg $\Delta$} \\
        \cmidrule{1-6}
        CT & WT & CT & WT & CT (2015) & WT (2019) & & & \\
        \midrule
        \hg\multicolumn{6}{c|}{\textbf{Human Study CT \cite{carlin2018political}}} & \textbf{0.65} & \textbf{1.22} & \textbf{0.93} \\
        \midrule 
        \hy \checkmark &  & \checkmark &  & \checkmark &  & 0.50 & 1.37 & 0.94 \\ 
        \checkmark &  & \checkmark &  &  & \checkmark & 1.26 & 0.81 & 1.04 \\ 
        \checkmark &  &  & \checkmark & \checkmark &  & 1.19 & 1.65 & 1.42 \\ 
        \checkmark &  &  & \checkmark &  & \checkmark & 1.21 & 2.08 & 1.65 \\ 
         & \checkmark & \checkmark &  & \checkmark &  & 1.01 & 1.60 & 1.31 \\ 
         & \checkmark & \checkmark &  &  & \checkmark & 1.31 & 1.62 & 1.47 \\ 
         & \checkmark &  & \checkmark & \checkmark &  & 1.42 & 1.79 & 1.61 \\ 
        \hy & \checkmark &  & \checkmark &  & \checkmark & 1.49 & 2.00 & 1.75 \\ 
        \midrule
        \hg\multicolumn{6}{c|}{\textbf{Human Study WT \cite{whitt2021tribalism}}} & \textbf{2.11} & \textbf{1.76} & \textbf{1.94} \\
        \bottomrule
        \end{tabular}
    }
    \vspace{-5pt}
\end{table*}
\subsection{Ablation: Effects of Backstories}
To further understand the effects of various parts of our
approach, we first consider the effect of extended backstories. Without the backstories, each virtual participant representation reduces to their answers to the demographic questions, which corresponds to the \BASEQA method. As we have previously discussed, \OURS substantially outperforms \BASEQA.

\subsection{Ablation Study: Effects of \DATE and \CONSISTENCY}

 We adopt the same experimental protocol as in our main experiments, focusing on Mistral-Small (24B) to isolate architectural factors.

\vspace{-3pt}
\paragraph{Dictator Game.}
In both the 2014~\citep{iyengar2015fear} and 2019~\citep{whitt2021tribalism} replications of the dictator game (see \Cref{tab:main_results_dictator_game}), we observe that the addition of either \DATE or \CONSISTENCY improves alignment with human partisan gaps compared to the base condition without either technique. Notably, combining both \DATE and \CONSISTENCY achieves the best performance. For the 2014 replication, this configuration yields a Mean Diff.\ $\Delta$ of 0.72, close to the human baseline of 0.68. Similarly, in the 2019~\citep{whitt2021tribalism} replication, the combined setup achieves a Mean Diff.\ $\Delta$ of 2.38—closely approximating the human gap of 2.40. These results suggest that providing models with both temporal context (via \DATE) and consistency enforcement improves their ability to reproduce partisan favoritism.

\paragraph{Trust Game.}
A similar trend is observed in the trust game simulations (\Cref{tab:main_results_trust_game}). Across both the 2015~\citep{carlin2018political} and 2019~\citep{whitt2021tribalism} human studies, models prompted with both \DATE and \CONSISTENCY yield the most faithful replications of human partisan bias. In the 2015 setting~\citep{carlin2018political}, the combined prompt produces a Mean Diff.\ $\Delta$ of 0.94, closely matching the human baseline of 0.93. In the 2019 setting~\citep{whitt2021tribalism}, the same configuration achieves a Mean Diff.\ $\Delta$ of 1.75—again nearly aligning with the human value of 1.94. Interestingly, \CONSISTENCY alone provides better results than \DATE alone in some cases (e.g., increasing the Rep $\Delta$ from 0.05 to 0.61 in 2015), highlighting the importance of both in reducing simulation bias.
\vspace{-3pt}
\subsection{Exploring Experiment Reproducibility}
\label{sec:axelrod_anova_experiment}
Reproducibility of human studies is a challenge in the social sciences. 
The present studies are typical, and are potentially influenced by 
{\bf Date} (Year of study), {\bf Framing} (wording of study instrction) and {\bf Participant Pools}. An intriguing and entirely new possibility enabled by LLM simulation it to counterfactually explore the effects of those factors on the experiment's outcomes. 

\subsubsection*{Date, Framing and Participant Pool Effects}
\label{sec:axelrod_framing_effect}

The two Dictator Game studies \cite{iyengar2015fear,whitt2021tribalism}, 
differed in year, framing (text of participant instructions, see \Cref{asec:study_instructions}), and participant pools.  Along with the two actually-occurring combinations of these factors in the studies, we simulate six more counterfactual combinations per study to produce a complete $2 \times 2 \times 2$ table of factor combinations. 
The results are shown in \Cref{table:axelrod_framing_effects}.
The orange rows represent the reproductions of the original human experiments, while counterfactual combinations occur between them with check marks indicating the factor combinations. The actual human study results are shown adjacent to the study replications, and  show good agreement with them.

The counterfactual rows allow us to compute main effects for each 
of the 3 contextual factors. e.g. from \Cref{table:axelrod_framing_effects} we can see in the Dictator game simulations, the main effects for participant pool, framing, and year 2014$\rightarrow$2019, are respectively -0.02, 1.18 and 0.4.

For the Trust game simulations, the main effects for participant pool, framing, and year 2015$\rightarrow$2019, are respectively 0.27, 0.41 and 0.16.

Note that in both cases, the main effects of participant instruction framing are stronger than the effects of year, in contrast with the conjectures of the authors of \citep{whitt2021tribalism}. The study instructions are given in \Cref{asec:study_instructions}, and it is noticeable that the Whitt study instructions contained boldface reminders of the partner's political affiliation at the end of the prompt. Recency bias, where a participants is more heavily influenced by more recently presented information, is a commonly cited factor in framing effects. 

It is also noticeable that the Whitt et al. Trust study instructions include an asymmetric (biased) description of possible outcomes, over-emphasizing the negative outcomes. 

We notice also a fairly strong participant pool effect (0.27) in the Trust game simulation, but not (-0.02) in the Dictator game. Examination of the respective research papers show that \citep{whitt2021tribalism} recruited participants via the online survey firm, Dynata. The earlier Dictator study~\citep{carlin2018political} used the online survey company SSI (Survey Sampling International). However, SSI and Dynata merged soon after, and appear to subsequently use similar recruiting methods and pools by the time of \citep{whitt2021tribalism}. It is not surprising then to see almost no participant effect between those studies. 
The earlier Trust study~\cite{iyengar2015fear} however used online participants recruited using Mechanical Turk, which has been shown to have more biased participant pools~\cite{PeerEtAl2022}.

More detailed analyses of the estimated effects, together with their
associated significance levels, are reported in
\Cref{asec:anova_results}.
The results indicate that, as expected, co-partisanship has a strong and significant effect on allocation. 

From \Cref{asec:anova_results}, we also see that allocation {\em difference} $\Delta$ is most strongly influenced by instruction framing (measured by {\em Same-Party} $\times$ {\em Framing} interaction) which is large and significant for both studies. 

On the other hand, study year affects allocation difference $\Delta$ significantly only in the Dictator studies.

Self-party affiliation affects $\Delta$ significantly in
the Trust Game but not in the Dictator Game. This is particularly important for studies which include unbalanced partisan pools, such as \citet{carlin2018political}.

Finally, the participant pools for the Trust studies (Dynata vs Mechanical Turk) had a weakly significant effect on $\Delta$.

Overall, these results suggest that while 
\citet{whitt2021tribalism} conjectured that increases in partisanship relative to earlier studies were due primarily to year of study, the present results suggest a more complex combination of question framing effects, study year and participant pool differences. 

\vspace{-5pt}
\section{Conclusion}
\vspace{-5pt}
We presented a framework for simulating complex human social behavior with language models.
Key to accurate simulation was the use of unbiased pretrained models with consistency filtering, followed by deep binding with richly constructed backstories derived from life-narrative interviews. 

We presented simulation studies with good agreement with earlier human studies, along with counter-factual studies that explored the disparities between human studies of the same phenomenon. 
While we agree that human studies should remain ``gold standard'' for behavioral studies, we argue that this work highlights often-neglected challenges of reproducibility in terms of study timing, participant pool and question framing effects, and can serve as a useful tool for informing the design and interpretation of future human studies.

\section*{Limitations}

\paragraph{Bias and Representational Limitations.}  
The distribution of generated virtual personas is constrained by the training data of the base LLMs.  
To the extent that these models reflect underlying biases—social, political, or cultural—those biases may propagate into the personas we generate.  
Even in the absence of explicit under-representation, matching-based approaches tend to yield less reliable estimates for minority groups, whose presence in the feature space is typically more sparse.  
This introduces a risk of larger errors for marginalized or ideologically atypical populations.

\paragraph{Data-Dependent Fidelity of Simulation.}  
The present paper and our prior work has shown promising fit to patterns of human behavior. That fidelity depends on those patterns being adequately represented in the model's training data. For example, our studies showed fidelity to participants bound to years 2015-2021. Many training documents containing contemporaneous dialog were crawled in those years. We would not expect similar fidelity if we attempted to bind participants to 1950. 

Similarly, there is a wealth of training data capturing users' self-experienced emotions, thoughts, and action descriptions which we expect can be accurately modeled. But a much more limited amount of data on the user's physiological or latent psychological state. We therefore don't expect to be able to model physiology, or unconscious psychological state. The issue of long-term behavior change is currently open. Individual dialog examples are short term, but the web does contain a non-trivial amount of ``diary'' type data from users, which allows us to model long-term effects. This is current work.

\paragraph{Resource Requirements.}  
Our approach depends on large models with extended context windows, coupled with compute-intensive generation and filtering pipelines.  
Generating high-quality backstories at scale—especially with rejection sampling and critic filtering—requires significant computational resources, which may pose barriers for smaller research groups or real-time applications.

\paragraph{Domain Specificity.}  
Our current experiments focus exclusively on U.S. political partisanship, reflecting the design of the human studies we replicate.  
As such, our findings may not generalize to other national, linguistic, or cultural contexts.  
Future work should explore broader applications, particularly in multilingual or non-Western settings.

\section*{Ethics Statement}

Our work aims to simulate aggregate-level human behavior using language models, not to anthropomorphize or personalize LLMs for individual interaction~\citep{cheng2024anthroscore}.  
Persona conditioning enables models to better approximate opinion distributions and ingroup biases, and also to better represent the diversity of the human population compared to directly steerable aggregate models. 
We caution against interpreting generated personas as real individuals or deploying them in contexts where anthropomorphic framing could lead to user misinterpretation.

A critical component of our methodology is the use of {\em pretrained} models that have not undergone fine-tuning for helpfulness, instruction-following and positivity.  
This is intentional: pretraining preserves a wider range of human-like expression, including disagreement, discomfort, and prejudice.  
As a result, some responses may reflect harmful biases, stereotyping, or negative emotion—just as real humans might in similar settings.  
By design, we want to be able to reproduce the kinds of dialog that arise in un-moderated social media platforms, as well as more regulated venues. 
We therefore urge caution in reading and sharing these responses, especially in public or sensitive domains.

Prior research has documented that LLMs may flatten human diversity, reinforce social stereotypes, or exaggerate partisan divisions—particularly when simulating marginalized groups~\citep{cheng2023marked, cheng2023compost, bai2024measuring, ostrow2025llms, wang2025large}.  
We recognize these risks and emphasize that our design choices aim to foreground such biases, not obscure them.

Finally, this work is intended solely for research purposes—especially within the social sciences.  
We strongly discourage its use for persuasive messaging, behavioral targeting, disinformation, or the impersonation of real individuals.  
Responsible deployment of virtual personas requires careful institutional oversight and transparent safeguards against misuse.

\section*{Acknowledgments}
J.S. and S.M. would like to acknowledge the support from the Korea Foundation for Advanced Studies (KFAS).
S.M. is supported by BAIR-Google Commons and M.K. is supported by the Apple Ph.D. Fellowship in Integrated Systems.

\bibliography{axelrod}

\clearpage

\appendix
\section{Study Instructions}
\label{asec:study_instructions}
Here are the two versions of the Dictator study instructions:

\tcbset{
  mybox/.style={
    enhanced,
    colback=white,       
    colframe=black,      
    boxrule=0.8pt,       
    arc=2mm,             
    left=6pt, right=6pt, top=6pt, bottom=6pt,
    colbacktitle=black,  
    coltitle=white,      
    fonttitle=\bfseries, 
    title=Dictator Game (Iyengar \& Westwood. 2015)
  }
}

\begin{tcolorbox}[mybox,width=0.5\textwidth]
\small{
\textbf{Question:} This game is played by pairs of individuals. Each pair is made up of a Player 1 and a Player 2. Each player will have some information about the other player, but you will not be told who the other players are during or after the experiment.
\medskip

The game is conducted as follows: A sum of \(\$10\) will be provisionally allocated to Player 1 at the start of each round. Player 1 will then decide how much of the \(\$10\)  to offer to player 2. Player 1 could give some, all, or none of the \(\$10\). Player 1 keeps all money not given to player 2. Player 2 gets to keep all the money Player 1 offers.
\medskip

The computer has randomly assigned you to play as Player 1. \textbf{You will play this game once with another person. The more money you obtain, the more successful you will be.}

\medskip 

You are Player 1. You have  \(\$10\). You can split the money between yourself and Player 2 in any way you want.

\medskip
Information about the Player 2

- Political party: \{Republican/Democrat\}

\medskip  
So put the dollars you wish to go to Player 2.
}
\end{tcolorbox}


\tcbset{
  mybox/.style={
    enhanced,
    colback=white,       
    colframe=black,      
    boxrule=0.8pt,       
    arc=2mm,             
    left=6pt, right=6pt, top=6pt, bottom=6pt,
    colbacktitle=black,  
    coltitle=white,      
    fonttitle=\bfseries, 
    title=Dictator Game (Whitt et al. 2021)
  }
}

\begin{tcolorbox}[mybox,width=0.5\textwidth]
\small{
\textbf{Question:} In this task you have \$10. You must decide how much to keep for yourself and how much to give to another person. In this task, the only information we can give you is that the other person identifies politically as a \{Republican/Democrat\}. Whatever you send to the other person will be given to them if they are selected to receive a payment. Whatever you keep for yourself will be paid to you if you are selected to receive a payment. Please make your decision.

\medskip  
How much money do you want to send to \{REPUBLICAN/DEMOCRAT\}? 
}
\end{tcolorbox}

Here are the two versions of the Trust study instructions:
\vspace{3.0in}
\eject
\tcbset{
  mybox/.style={
    enhanced,
    colback=white,       
    colframe=black,      
    boxrule=0.8pt,       
    arc=2mm,             
    left=6pt, right=6pt, top=6pt, bottom=6pt,
    colbacktitle=black,  
    coltitle=white,      
    fonttitle=\bfseries, 
    title=Trust Game (Carlin \& Love 2016)
  }
}

\begin{tcolorbox}[mybox,width=0.5\textwidth]
\small{
\textbf{Question:}  In this game, there are two players: “Player 1” and “Player 2.” You are Player 1.
\medskip

Player 2, like you, received 10 raffle tickets to participate in the study. You will not be told who this person is either during or after the survey (nor will that person be told who you are). We have chosen to give you one piece of information about Player 2:
\medskip

\textbf{Player 2 identifies politically with the \{Democratic/Republican\} Party.}

\medskip
Today, you will have the opportunity to send some, all, or none of 10 raffle tickets to Player 2. This person will then keep the tickets sent to them, if any, and you keep the remainder.
\medskip

The game is as follows:
\begin{itemize}[leftmargin=2em]   
\vspace{-1em}
    \item[1.)] You must decide how many raffle tickets (if any) you will send to Player 2. Any amount you keep for yourself will be given to you.
    \vspace{-0.5em}
    \item[2.)] Whatever amount you send to Player 2 will be tripled before it is passed on to Player 2. Player 2 then has the option of returning any portion of this tripled amount to you. Then the game is over.

\end{itemize}
\textbf{
For example:
\begin{itemize}[leftmargin=2em] 
\vspace{-1em}
    \item[(a)] If you send 1 ticket, Player 2 receives 3 tickets.
    \vspace{-0.5em}
    \item[(b)] If you send 5 tickets, Player 2 receives 15 tickets.
\end{itemize}
\vspace{-0.5em}
}
These are examples only.
To review, any number of tickets you send will be tripled and given to Player 2. This person will keep the tickets you send them, minus the number of tickets they return to you. Your payoff in this game is the number of tickets that you kept for yourself, plus any tickets Player 2 returned to you.
Once you understand this game, and are ready to make a decision, select the amount from the options below.

}
\end{tcolorbox}

\tcbset{
  mybox/.style={
    enhanced,
    colback=white,       
    colframe=black,      
    boxrule=0.8pt,       
    arc=2mm,             
    left=6pt, right=6pt, top=6pt, bottom=6pt,
    colbacktitle=black,  
    coltitle=white,      
    fonttitle=\bfseries, 
    title=Trust Game (Whitt et al. 2021)
  }
}

\begin{tcolorbox}[mybox,width=0.5\textwidth]
\small{
\textbf{Question:} In this task you have \(\$10\). You must decide how much to keep for yourself and how much to give to another person. The only information we can give you is that the other person identifies politically as a \{Republican/Democrat\}.  

\medskip

Whatever you keep for yourself will be paid to you if you are selected to receive a payment. Whatever you send to the other person will be multiplied by 3 and then given to the other person. The other person then has the option to give money back to you.  

\medskip

\textbf{For example:}
\vspace{-0.5em}
\begin{itemize}[leftmargin=0.5em]
    \item If you keep \(\$10\), then the other person receives \(\$0\).
    \vspace{-0.5em}
    \item If you send \(\$10\), then we multiply that sum by 3 \((\$10 \times 3 = \$30)\) and the other person receives \(\$30\).
    \vspace{-0.5em}
    \item The other person then decides how much (if any) of the \(\$30\) to give back to you.
    \begin{itemize}
        \vspace{-0.5em}
        \item If the person keeps all \(\$30\), then you receive \(\$0\).
        \item If the person returns half, then you and the other person receive \(\$15\) each.
        \vspace{-0.5em}
    \end{itemize}
\end{itemize}

How much money do you want to send to \{REPUBLICAN/DEMOCRAT\}? 
}
\end{tcolorbox}
\begin{table*}[t]
\centering
\caption{
\textbf{Estimated Main Effects and Same-Party interactions for human Dictator and Trust Studies.}
Entries report coefficient estimates from OLS factorial models.
Statistical significance of non-zeros coefficient is indicated by asterisks ($^{***}p<0.001$, $^{**}p<0.01$, $^{*}p<0.05$).
}
\label{tab:human_effects}
\resizebox{4.4in}{!}{
\begin{tabular}{lccccc}
\toprule
& \multicolumn{3}{c}{Main Effects} & \multicolumn{2}{c}{SameP interactions}\\
\cmidrule(lr){2-4} \cmidrule(lr){5-6}
& SameP & SelfP & Study & $\times$SelfP &  $\times$Study\\
\midrule
{\bf Dictator Game (ID vs.\ WD)} &
0.577*** &
0.635*** &
-0.259 &
0.047 & 
1.801*** \\
\addlinespace
{\bf Trust Game (CT vs.\ WT)} &
0.739* &
-0.182 &
-0.466 & 
0.434 &
0.973**
 \\
\bottomrule
\end{tabular}
}
\end{table*}
\begin{table*}[t]
\centering
\caption{
\textbf{Estimated Main Effects and Same-Party interactions for counterfactual simulations of the Dictator and Trust Games.}
Entries report coefficient estimates from linear models fit to the full factorial set of simulated counterfactual combinations.
Statistical significance for non-zero coefficients is indicated by asterisks ($^{***}p<0.001$, $^{**}p<0.01$, $^{*}p<0.05$).
}
\label{tab:simulation_effects_combined}
\resizebox{\textwidth}{!}{
\begin{tabular}{lccccccccc}
\toprule
& \multicolumn{5}{c}{Main Effects} &
\multicolumn{4}{c}{SameP Interactions}\\
\cmidrule(lr){2-6} \cmidrule(lr){7-10}
& SameP & SelfP & Year & Framing & Pool & $\times$SelfP & $\times$Year & $\times$Framing & $\times$Pool\\
\midrule
\textbf{Simulations (Dictator)}
& 1.464*** 
& 0.243***
& 0.073
& -0.314***
& -0.046
& -0.145 
& 0.416*** 
& 1.134*** 
& -0.009 \\
\addlinespace
\textbf{Simulations (Trust)}
& 1.369*** 
& 0.435***
& 0.159*
& 0.129
& 0.105
& -0.474***
& 0.195 
& 0.366**
& 0.324* \\
\bottomrule
\end{tabular}

}
\end{table*}

\section{Regression analysis of Human Studies and Counterfactual Simulations}
\label{asec:anova_results}

\paragraph{Notation and study labels.}
We refer to the two human Dictator Game studies as
\textbf{ID} \citep{iyengar2015fear} and \textbf{WD} \citep{whitt2021tribalism},
and to the two Trust Game studies as
\textbf{CT} \citep{carlin2018political} and \textbf{WT} \citep{whitt2021tribalism}.

\paragraph{Model specification and variables (human studies).}
Each participant plays either a Trust or Dictator game with either a same-party or a different-party partner. The participant $i$ response variable is their allocation
$y_i$ to their partner.

We model their allocation using a factor model with factors:
$\textit{SameP}_i = 1$ if the partner is a co-partisan, 0 otherwise. 
We define $\textit{SelfP}_i = 1$ for Democratic allocators and $0$ for Republican allocators;
and $\textit{Study}_i = 1$ for observations drawn from the Whitt studies
(\textbf{WD} in the Dictator Game and \textbf{WT} in the Trust Game), with
$\textit{Study}_i = 0$ denoting the corresponding Iyengar
(\textbf{ID}) or Carlin (\textbf{CT}) studies.

Note that there is no direct measurement of partisan allocation difference in this between-participants design. The difference is instead a modeled variable. It is the main effect of the \textit{SameP} factor.
To measure the effect of other variables on the difference, we must consider {\em interactions} with the \textit{SameP} factor.

The Dictator Game sample includes $N_{\textbf{ID}} = 1{,}711$ observations from
\textbf{ID} (888 Democrats and 823 Republicans) and
$N_{\textbf{WD}} = 801$ observations from \textbf{WD}
(444 Democrats and 357 Republicans),
for a total of $N = 2{,}512$ Dictator Game observations.

The Trust Game sample includes $N_{\textbf{CT}} = 576$ observations from
\textbf{CT} (388 Democrats and 188 Republicans) and
$N_{\textbf{WT}} = 801$ observations from \textbf{WT}
(444 Democrats and 357 Republicans),
for a total of $N = 1{,}377$ Trust Game observations.

For each game, we estimate the regression model:
\begin{equation}
\label{eq:human_ols}
\begin{aligned}
y_{ijk}~=~&\beta_0 + \\
&\beta_1 \textit{SameP}_i + \\
&\beta_2 \textit{SelfP}_j +\\
&\beta_3 \textit{Study}_k +\\
&\beta_4 \textit{SameP}_i \times \textit{SelfP}_j + \\
&\beta_5 \textit{SameP}_i \times \textit{Study}_k + \\
&\varepsilon_{ijk}
\end{aligned}
\end{equation}
where $y_{ijk}$ is the mean response for the ${ijk}$ participant group, and $\epsilon_{ijk}$ is a zero-mean error term.

\paragraph{Human Regression results.}
\Cref{tab:human_effects} reports coefficient estimates and significance for non-zero coefficients.
First, recall that the response variable here is the allocation, not the partisan difference. The first column shows a significant positive main effect of {\em SameP} which is that participants allocate more to same-party peers. The self-party allocation was somewhat higher for Democrats (second column) in the Dictator game, but not for the Trust game. 

Turning to the interactions, we see that high {\em SameP} effects are predicted by the Study factor (last column), with the later study (Study = 1) corresponding to significanly higher partisan allocation difference. 

\paragraph{Interpretation of human regression effects.}

The human study results validate the
respective papers' hypotheses: i.e. there are significant positive main effects on allocation for same-party peers (first column), and that the
later studies predict higher partisan
allocation differences (last column). 

\paragraph{Motivation for counterfactual simulations.}
Because the $\textit{Study}$ indicator aggregates multiple contextual changes—
including historical timing, instructional framing, and participant pool—the human
data alone cannot identify which dimensions drive changes in partisan favoritism.
To disentangle these mechanisms, we conduct counterfactual simulations that
independently recombine $\textit{Year}$, $\textit{Framing}$, and $\textit{Pool}$
while holding the game structure fixed.

\paragraph{Model specification (counterfactual simulations).}
The counterfactual simulations are constructed as factorial
experiments over an expanded set of conditions. In addition, since it is
easy to repeat experiments with the same
virtual participant without order effects, we
place each virtual participant in the allowable counter-factual states. That is, we vary {\em SameP}, {\em Year} and {\em Framing} for each participant. This gives us $8\times$ as many measurements as the original human studies. We do not alter the participant's self-party or participant pool. In the first case, the participant's political affiliation is a fixed feature of their identity. In the second case, the effect of participant pool (which we are interested in measuring) is only defined by the members in it, and altering membership would erase pool effects. 
 
Note that now there is a total
of 5 independent variables,
3 of which: 
\textit{SameP}, \textit{Year}, and \textit{Framing}, vary within-participant, and 2 of which:
\textit{SelfP} and \textit{Pool}, are between participant variables.
When computing main effects or interactions, we take care to aggregate first for each participant to obtain independent measures, and then apply an appropriate one-sided (for the within-participant effects) or two-sided (for between participant effects) test. 

The motivation for this factorial design is to isolate which contextual dimensions
drive variation in the strength of same-party favoritism.
In the human data, these dimensions are bundled within $\textit{Study}$ and cannot
be independently manipulated.
By contrast, the simulation framework allows us to recombine $\textit{Year}$,
$\textit{Framing}$, and $\textit{Pool}$ while holding the strategic structure of the games fixed.

Let $y_{ijk\ell m}$ denote the allocation by a virtual user for a factor combination comprising co-partisanship indicator ($i$) allocator party ($j$), study year
($k$), framing ($\ell$), and participant pool ($m$).
Binary indicators are oriented as follows:
$\textit{SameP} = 1$ for co-partisans, 0 for anti-partisans. 
$\textit{SelfP} = 1$ denotes a Democratic allocator.
Other contextual indicators are oriented relative to the Whitt studies.
Specifically, $\textit{Year} = 1$ corresponds to the calendar year of the Whitt
experiments (\textbf{WD} or \textbf{WT}),
$\textit{Framing} = 1$ corresponds to the game instructions used in the Whitt
studies, and $\textit{Pool} = 1$ corresponds to the Whitt participant pool.
A value of $0$ denotes the corresponding dimension taken from the Iyengar
(Dictator Game) or Carlin (Trust Game) studies.

To analyze this factorial experiment, we estimate a linear model once again.
Formally, we estimate the following OLS model:
\begin{equation}
\label{eq:simulation_ols}
\begin{aligned}
y_{ijk\ell m}
=&\;
\alpha_0 + \\
&\alpha_1 \textit{SameP}_{i} + \\
&\alpha_2 \textit{SelfP}_{j} + \\
&\alpha_3 \textit{Year}_{k} + \\
&\alpha_4 \textit{Framing}_{\ell} + \\
&\alpha_5 \textit{Pool}_m  +\\
&\alpha_6\textit{SameP}_{i} \times \textit{SelfP}_j + \\
&\alpha_7\textit{SameP}_{i} \times \textit{Year}_k + \\
&\alpha_8\textit{SameP}_{i} \times \textit{Framing}_{\ell} + \\
&\alpha_9\textit{SameP}_{i} \times \textit{Pool}_{m} + \\
&\epsilon_{ijk\ell}
\end{aligned}
\end{equation}
where $u_{ijk\ell}$ is a mean-zero error term.

\paragraph{Simulation results.}
\Cref{tab:simulation_effects_combined} reports coefficient estimates from the
counterfactual simulations.

First of all, {\em SameP} main effects are large, indicating a strong average allocation difference between co-partisans and anti-partisans. 

Secondly, {\em SelfP} main effects are moderately strong, indicating a higher average allocation for Democrats. 

Framing had a main effect on overall allocation in the Dictator games only. 

For the factors that affect allocation difference ({\em SameP} interactions):

{\em SameP}$\times${\em SelfP} effects are modest and indicate a significant partisan effect for Republicans in the Trust study only. 

{\em SameP}$\times${\em Year} effects are modest, and significant only for the Dictator simulation. 

{\em SameP}$\times${\em Framing} effects are strong, and significant for both Dictator and Trust simulations. As discussed in the main text, there are several features in the wording of the later studies that normally cause strong framing effects. 

Finally, {\em SameP}$\times${\em Pool} effects are relatively weak, and almost vanish for the Dictator studies, which appear to have used the same recruiting service for their participant pools.  However, participant pool is weakly significant for the Trust studies, which included one pool from Dynata/SSI and one pool from Mechanical Turk. 
\section{Backstory Generation}
\label{asec:backstories}
To generate high-quality virtual backstories, we follow the backstory construction pipeline introduced by~\citet{kang2025higher}.  
We adopt the same set of interview questions used in their work to elicit long-form, coherent, and persona-consistent narratives.  
\Cref{table:appendix_list_of_interview_questions} provides the full list of prompts used to construct these interview-style backstories.  

Backstories are generated iteratively by prompting a language model with each question in sequence, along with all previously generated question–answer pairs.  
Specifically, to generate a response for the $i$-th question, the model is conditioned on the current prompt and the full context from steps $1$ through $(i-1)$.  
This incremental setup promotes temporal coherence and a consistent persona across the entire narrative.  
To further enhance quality, we apply rejection sampling guided by an LLM-as-a-critic model, which filters out low-quality generations.  
The resulting backstories are both expressive and internally consistent, enabling faithful simulation of human-like personas.

We provide the examples of generated virtual personas for Democrat and Republican virtual participants below:
\begin{tcolorbox}[width=1.0\linewidth, halign=left, colframe=black, colback=white, boxsep=0.01mm, arc=1.5mm, left=2mm, right=2mm, boxrule=1pt, breakable, enhanced]
\footnotesize{
\textbf{Example Interview-Style Backstory of a \emph{Democrat} User}\\
\vspace{10pt}
Question: To start, I would like to begin with a big question: tell me the story of your life. Start from the beginning--from your childhood, to education, to family and relationships, and to any major life events you may have had.
\vspace{5pt}
\newline Answer: I have always lived in Maryville. I am 18 years old and was born on March 31st, 1984. I grew up with my mother, father, and two younger brothers, Jacob and Zachary. My father works at a steel mill and my mother is a nurse at the hospital. 

\vspace{5pt}
My childhood was rather ordinary. I went to school, hung out with my friends, and don’t recall any major life events. I attended Maryville High School, completed the required courses with some electives, and graduated in 2003.  

\vspace{5pt}
I have never been in a serious relationship. I’ve had a few crushes, but nothing major. My current crush is my friend Alex—we’re very close but have never been romantic. At 18, I don’t want to be tied down. I’d rather enjoy life while I’m young.  

\vspace{5pt}
I’m currently a sophomore in college majoring in Psychology. I’m not sure yet what I’ll do with my degree; right now my focus is on getting through school and being happy.

\vspace{5pt}
Question: Some people tell us that they've reached a crossroads at some points in their life where multiple paths were available, and their choice then made a significant difference in defining who they are. What about you? Was there a moment like that for you, and if so, could you tell me the whole story about that from start to finish?
\vspace{5pt}\\
Answer: I’ve never had a dramatic crossroads or done anything particularly exciting. The one decision that shaped my life was choosing to go to college. Before that, I wasn’t sure what I wanted to do. Going to college gave me direction and made me feel like I was on the right track. I’m doing well in school and looking forward to the future.

\vspace{5pt}
Question: Tell me about anyone else in your life we haven’t discussed (like friends or romantic partners). Are there people outside of your family who are important to you?
\vspace{5pt}\\
Answer: My best friend is Alex, the one I mentioned before. We spend a lot of time together and I really value his company. I have a few other friends I see occasionally, but Alex is the most important. I don’t have any romantic partners right now and I’m fine with that. I feel lucky to have a good family, good friends, and a good life.

\vspace{5pt}
Question: Now let’s talk about your current neighborhood. Tell me all about the neighborhood and area in which you are living now.
\vspace{5pt}
\newline Answer: I live in a small town called Maryville, about an hour from Asheville. It’s a calm area with lots of trees and parks. I enjoy the slower pace of life here and the ability to walk outside and enjoy the fresh air.  

\vspace{5pt}
Maryville has a few bars and restaurants but not much else going on, which I actually like. I prefer small towns over big cities. One downside is the lack of job opportunities—most people work at the steel mill, attend school, or commute elsewhere. Finding well-paying work here is hard, which makes moving out on my own challenging. Still, overall I’m happy living here.

\vspace{5pt}
Question: Tell me about any recent changes to your daily routine.
\vspace{5pt}
\newline Answer: Not much has changed. My days are already busy with school, work, and spending time with friends. I’m also trying to make more time for my art. My routine helps me manage stress: I wake up early, do chores, work until noon, then cook lunch. In the afternoon I go to school or work, and in the evening I study until about 7 pm. After that, I have free time—usually for my art or hanging out with friends. Adding more leisure activities would help me feel less stressed, but overall the structure keeps me balanced.

\vspace{5pt}
Question: How would you describe your political views?
\vspace{5pt}
\newline Answer: I’m not very interested in politics. I usually go along with what my parents—who are Democrats—believe. I think it’s important to have some kind of government, and I’m grateful for the opportunities I have in this country, but politics isn’t a major focus in my life.

\vspace{5pt}
Question: How have you been thinking about race in the U.S. recently?
\vspace{5pt}
\newline Answer: I don’t think about race very much. I’ve always had friends of different races and believe everyone should be treated equally. I dislike racist jokes and think it’s ignorant when people feel superior because of their race. In my view, race shouldn’t be a barrier to success in this country.  

\vspace{5pt}
Question: For you, what makes it easy or hard to stay healthy?
\vspace{5pt}\\ 
Answer: Staying healthy isn’t too difficult for me. I’m fairly active and try to eat well. I’ve never had major health problems, though I sometimes feel self-conscious about my weight since I’m tall and heavier than average. Being active helps me feel healthy and confident.

\vspace{5pt}
Question: Some people are excited about medical vaccination, and others, not so much. How about you?
\vspace{5pt}\\ 
Answer: I’m not entirely sure how I feel about vaccinations. I know they’re important for preventing illness—I’m glad I was vaccinated as a child—but I have friends who think they’re unsafe. I don’t believe vaccines are dangerous, but I’m not fully convinced either. I think vaccination should be a personal choice rather than a requirement, though more research could help clarify safety concerns.

\vspace{5pt}
Question: Some people say they struggle with depression, anxiety, or something else like that. How about for you?
\vspace{5pt}\\ 
Answer: I think I might have some depression or something similar. Sometimes I feel really down and sad without knowing why. Stress causes headaches and a racing heart, and sometimes I feel like I might have a panic attack. Talking to people helps a little, but most of the time I prefer to be alone. I’m not sure how to make myself feel better, but I try to manage it as best as I can.
}
\end{tcolorbox}
\begin{tcolorbox}[width=1.0\linewidth, halign=left, colframe=black, colback=white, boxsep=0.01mm, arc=1.5mm, left=2mm, right=2mm, boxrule=1pt, breakable, enhanced]
\footnotesize{
\textbf{Example Interview-Style Backstory of a \emph{Republican} User}\\
\vspace{10pt}
Question: To start, I would like to begin with a big question: tell me the story of your life. Start from the beginning--from your childhood, to education, to family and relationships, and to any major life events you may have had.
\vspace{5pt}
\newline Answer: Sure, I was born in the year 2000 and was raised by a single mother in a small town outside of Seattle. Growing up, my mother worked hard to provide for me and I had to rely heavily on my innate curiosity and self-motivation to pursue my educational and career objectives.

\vspace{5pt}
As a child, I always loved gadgets and technology, so I built my first computer when I was just 11 years old. Around that same time, I became interested in aviation and spent hours building and flying model airplanes. When I was 13, I enrolled in an online aviation course and flew my first real airplane at the age of 14. I was hooked.

\vspace{5pt}
Question: Some people tell us that they've reached a crossroads at some points in their life where multiple paths were available, and their choice then made a significant difference in defining who they are. What about you? Was there a moment like that for you, and if so, could you tell me the whole story about that from start to finish?
\vspace{5pt}\\
Answer: Sure, I took the red pill when I decided to pursue a career in computer science and artificial intelligence. The choice wasn't necessarily an easy one for me since I had always been intrigued by aviation and even had my pilot's license at a young age. However, after taking a year off between high school and college, I realized that my true passion lies in computer science and artificial intelligence and decided to pursue it wholeheartedly.
\vspace{5pt}\\
This decision has shaped my professional path, leading me to get a bachelor's degree in computer science from a prestigious university and work on cutting-edge projects on machine learning and NLP. It also led me to start my own business providing AI consulting services to clients worldwide. The choice to pursue AI has been one of the best decisions of my life, and I have enjoyed every moment of my career journey so far.

\vspace{5pt}
Question: Tell me about anyone else in your life we haven’t discussed (like friends or romantic partners). Are there people outside of your family who are important to you?
\vspace{5pt}\\
Answer: Sure, I have a group of close friends whom I met during college. We all had a similar passion for technology, and we bonded over late-night study sessions and group projects. We support each other's passions and goals and help each other navigate challenges while pursing our dreams. I also have a fiancé whom I met while working on an AI project. She is also a software engineer and has been instrumental in shaping my career decisions. Together, we work on building machine learning models and experimenting with new libraries and tools. She has been an amazing partner, and I cannot imagine my life without her support and love.

\vspace{5pt}
Question: Now let’s talk about your current neighborhood. Tell me all about the neighborhood and area in which you are living now.
\vspace{5pt}
\newline Answer: Currently, I live in a quiet neighborhood near downtown Seattle, Washington. The neighborhood is conveniently located near public transportation, making it easier for me to travel around the city without having to drive a car.  There are many parks nearby that offer opportunities for outdoor activities like hiking and biking. The area has good schools, and we are living in a classic 1920s Tudor house with a few modern amenities added.
\vspace{5pt}\\
The community is highly diverse and has a bustling cultural scene with local festivals, art galleries, music venues, and theaters.  The area also offers a great selection of shopping, dining, and entertainment opportunities, and there are a variety of local businesses that keep the community thriving.  Overall, it's a great place to live and work, and I feel lucky to be a part of it.

\vspace{5pt}
Question: Tell me about any recent changes to your daily routine.
\vspace{5pt}
\newline Answer: Sure, along with starting a new job this month,  I have recently started a new routine for exercising. Now I go for a brisk walk early in the morning before getting to work.  In addition, I have also started cooking dinner every night, which has led to me eating more homemade, healthy meals.
\vspace{5pt}\\
After a grueling day at work, I unwind by indulging in video games and creative writing or watching my favorite shows and movies on Netflix or Disney+.  Before bed, I also like to catch up on news and social media on my phone, and I usually read a chapter of a book before sleeping.  All in all, it's great to have a healthy routine in place and keep a balance between work, health, and leisure.  By incorporating these small changes, I feel more balanced and productive in my daily life.

\vspace{5pt}
Question: How would you describe your political views?
\vspace{5pt}
\newline Answer: I identify as a conservative and believe in protecting our nation's security through strong foreign policies. I am also in favor of lower taxes so as to stimulate the economy and job creation, while also considering the needs of less fortunate members of our community. As a Christian, I have a deep sense of empathy and compassion for those who are in need of assistance, which has instilled a sense of civic duty in me. Generally, I our political views should be fluid and evolving our society develops. Our party differences should not get in the way of achieving our shared goal of contributing to our nation's growth and prosperity for all Americans.

\vspace{5pt}
Question: How have you been thinking about race in the U.S. recently?
\vspace{5pt}
\newline Answer: It has become increasingly disheartening to witness racial injustice and police brutality against black Americans in our country. The protests and riots that have been sweeping across the nation are a natural response to the frustration that African Americans have endured for centuries.  I am encouraged by the strong support coming from both sides of the aisle and believe that by coming together, we can finally address the systemic injustices that have long plagued our community. I remain optimistic that meaningful change is possible if bipartisan efforts and solutions are implemented  to create more equitable societies for everyone within our country.  The journey to a more fair and just society is fraught with uncertainties and setbacks. But we should all be encouraged by the sincere efforts we see across our nation to build bridges and seek reforms. Together, we can create a better America for all.

\vspace{5pt}
Question: For you, what makes it easy or hard to stay healthy?
\vspace{5pt}\\ 
Answer: For me, one of the biggest factors in maintaining good health is managing stress.  The daily grind of work, family obligations, and maintaining a household can take their toll, making it difficult to prioritize health and wellness. Additionally, I have a busy work schedule, leaving less time for physical activities or meal preps, which can interfere with a healthy lifestyle.  I try to incorporate relaxation practices into my routine to help keep stress levels low and ensure that I have enough time to take my health seriously.  Other factors like diet, exercise, and sleep also play crucial roles; however, my focus has primarily been on managing stressors and finding ways to make the most of limited leisure time so that my health does not suffer.

\vspace{5pt}
Question: Some people are excited about medical vaccination, and others, not so much. How about you?
\vspace{5pt}\\ 
Answer: absolutely, I am very excited about medical vaccinations. I believe it is a critical step in preventing spread of contagious diseases. Vaccines are designed to stimulate the body's immune system to create antibodies that protect us from future infections from certain diseases.  They have already saved countless lives, especially children, from potentially deadly diseases like measles, rubella, Hib disease, etc.  So, yes, without a doubt, I think getting vaccinated against these harmful pathogens should be mandatory and taken very seriously by everyone, especially parents to safeguard their child's health.  In short, I am a strong advocate for medical vaccinations and encourage everyone around me to prioritize the health and safety of their children by getting them vaccinated.

\vspace{5pt}
Question: Some people say they struggle with depression, anxiety, or something else like that. How about for you?\
\vspace{5pt}\\ 
Answer: Of course, I have my moments like everyone else; however, I manage my emotions to the best of my abilities.  I have never been diagnosed with clinical depression or anxiety, but I don't shy away from seeking help when needed.
\vspace{5pt}\\
I have experienced some mental health challenges, such as following the death of my father, but in the long run, those were temporary setbacks.  As a coping mechanism, I try to stay grounded and mindful of my feelings by engaging in hobbies such as running, hiking, and exploring new places.  This practice has allowed me to manage my mental health while keeping stress levels low and enjoying life to its fullest.
}
\end{tcolorbox}

\begin{table*}[ht]
    \centering
    \scriptsize
    \caption{
        Abridged set of interview questions used to generate transcript-style backstories.  
        These prompts are adapted from oral history protocols developed by the American Voices Project~\citep{stanford2021american}, covering key themes such as family, education, work, health, politics, and community life.
    }

    \label{table:appendix_list_of_interview_questions}
    \resizebox{\textwidth}{!}{
    \begin{tabular}{p{0.05\textwidth}|p{0.95\textwidth}}
    \toprule
    \textbf{Q\#} & \textbf{Interview Question} \\
    \midrule
    1 & To start, I would like to begin with a big question: tell me the story of your life. Start from the beginning--from your childhood, to education, to family and relationships, and to any major life events you may have had. \\
    \midrule
    2 & Some people tell us that they've reached a crossroads at some points in their life where multiple paths were available, and their choice then made a significant difference in defining who they are. What about you? Was there a moment like that for you, and if so, could you tell me the whole story about that from start to finish? \\
    \midrule
    3 & Tell me about anyone else in your life we haven't discussed (like friends or romantic partners). Are there people outside of your family who are important to you? \\
    \midrule
    4 & Now let's talk about your current neighborhood. Tell me all about the neighborhood and area in which you are living now. \\
    \midrule
    5 & Tell me about any recent changes to your daily routine. \\
    \midrule
    6 & How would you describe your political views? \\
    \midrule
    7 & How have you been thinking about race in the U.S. recently? \\
    \midrule
    8 & For you, what makes it easy or hard to stay healthy? \\
    \midrule
    9 & Some people are excited about medical vaccination, and others, not so much. How about you?
    \\
    \midrule
    10 & Some people say they struggle with depression, anxiety, or something else like that. How about for you? \\
    \bottomrule
    \end{tabular}
    }
\end{table*}

\section{Demographic Survey}
After generating backstories through open-ended narrative sampling, we emulate the process of collecting sociodemographic and ideological information by administering a structured survey to each virtual persona~\citep{moon2024virtual,kang2025higher}.  
This step is essential for curating a pool of virtual personas whose aggregate trait distribution closely mirrors that of the human population we aim to simulate.  
The resulting demographic metadata is later used in the user–persona matching process described in \Cref{asec:demographic_matching}.

Unlike human respondents, who each possess a fixed and known set of demographic and ideological characteristics, virtual personas do not necessarily exhibit a unique or explicit trait profile.  
A single backstory may implicitly represent a range of possible individuals unless specific details are directly stated (e.g., ``I am a 30-year-old woman").  
As a result, we represent each persona’s traits using a probability distribution rather than a deterministic vector.

We adopt a two-stage procedure for constructing these distributions, following the approach of~\citet{moon2024virtual,kang2025higher}.  
In the first stage, we attempt to extract explicit trait mentions directly from the narrative.  
To do this, we prompt \texttt{gpt-4o} (with temperature \texttt{T = 0}) using a set of targeted trait-identification questions and the backstory as input.  
If the model finds clear verbal evidence for a given trait (e.g., ``I’m a proud Democrat"), we assign a one-hot distribution to that trait category.  
If no explicit mention is found, we proceed to the second stage, where the model infers a probability distribution over the possible trait values.  

Below, we provide the full list of trait-seeking prompts used to elicit evidence for six key sociodemographic and ideological attributes.
\section{Demographic Matching}
\label{asec:demographic_matching}

To approximate survey responses using synthetic personas, we assign each real survey participant to a virtual backstory that closely mirrors their demographic and ideological profile. We model this assignment process as a complete weighted bipartite graph \( G = (H, V, E) \), where \( H = \{h_1, \dots, h_n\} \) denotes the set of \( n \) human users, and \( V = \{v_1, \dots, v_m\} \) is a larger pool of \( m \) candidate virtual personas, with \( m > n \) to allow for flexible and diverse matching.

Each human user \( h_i \) is described by a fixed \( k \)-dimensional tuple of categorical traits, \((t_{i1}, \dots, t_{ik})\), spanning demographic and ideological characteristics. Each virtual persona \( v_j \), in contrast, is associated with a probability distribution over these \( k \) traits: 
\[
\left(P(d_{j1}), \ P(d_{j2}), \ \dots, \ P(d_{jk})\right)
\]
The weight of the edge \( e_{ij} \in E \), connecting human user \( h_i \) to persona \( v_j \), represents the likelihood that \( v_j \) matches all of \( h_i \)’s traits, and is defined as:
\begin{equation}
w(e_{ij}) = \prod_{l=1}^k P(d_{jl} = t_{il})
\end{equation}

We then apply a maximum-weight bipartite matching algorithm to find an injective mapping \( \pi: [n] \rightarrow [m] \) that maximizes the total matching score:
\[
\sum_{i=1}^n w(h_i, v_{\pi(i)})
\]
This optimization is solved exactly using the Hungarian algorithm~\citep{Kuhn1955Hungarian}. After matching, we select \( n \) virtual backstories whose traits collectively reflect the demographic distribution of the original human sample, ensuring that our synthetic population maintains fidelity to the target survey group.
\section{Experiment Details for Reproducibility}
\label{asec:experiment_details}

We run all experiments using either 8 NVIDIA RTX A6000 GPUs or 8 NVIDIA RTX A5000 GPUs, depending on availability.  
Interview-style backstories are generated from three pretrained language models:  
\texttt{Llama-2-70B}~\citep{Touvron2023LLaMA},  
\texttt{Llama-3.1-70B}~\citep{llama3},  
and \texttt{Mistral-Small-24B-Base-2501}~\citep{mistral2025small24b}.  
For all generation runs, we use a decoding temperature of \texttt{T = 1.0}.  

To filter low-quality generations, we employ \texttt{Gemini-2.0}~\citep{gemini2024} as a large language model (LLM) critic. The critic performs binary classification over each generated backstory using deterministic decoding (\texttt{T = 0}) and serves as the basis for rejection sampling.  

For inference and response likelihood estimation across all methods, we adopt offline batched decoding using vLLM version 0.10.2~\citep{kwon2023efficient}.  
All code, prompts, and generated backstories will be publicly released to support reproducibility.

\end{document}